\documentclass[runningheads]{llncs}

\usepackage{eccv}

\usepackage{eccvabbrv}

\usepackage{graphicx}
\usepackage{booktabs}

\usepackage{tikz}
\usetikzlibrary{arrows.meta,positioning}
\usepackage{pgfplots}
\pgfplotsset{compat=1.18} 

\usepackage[accsupp]{axessibility}  

\usepackage[pagebackref,breaklinks,colorlinks,citecolor=eccvblue]{hyperref}
\usepackage{hyperref}

\usepackage{orcidlink}

\begin{document}

\title{On-Orbit Real-Time Wildfire Detection Under On-Board Constraints} 

\titlerunning{Real-Time On-Orbit Wildfire Segmentation}

\author{Matthias R\"otzer\inst{1}\orcidlink{0009-0003-6894-5027}  \and
Veronika P\"ortge\inst{1}\orcidlink{0000-0001-9019-6294} \and
Martin Ickerott\inst{1} \and
Jayendra Praveen Kumar Chorapalli\inst{1} \and
Dimitri Scheftelowitsch\inst{1}\orcidlink{0000-0002-6650-4434} \and
Max Bereczky\inst{1}\orcidlink{0000-0002-4027-3295} \and
Dmitry Rashkovetsky\inst{1}\orcidlink{0000-0002-6650-4434} \and
Sai Manoj Appalla\inst{1} \and
Julia Gottfriedsen\inst{1,2}\orcidlink{0000-0002-6650-4434}}

\authorrunning{M. R\"otzer et al.}

\institute{OroraTech GmbH, St.-Martin-Str. 112, 81669 Munich, Germany \\
\and
Ludwig-Maximilians-Universität München, Munich, Germany}

\maketitle

\begin{abstract}
We present a deployed system for on-orbit wildfire detection aboard a nine-satellite commercial thermal infrared constellation (OroraTech OTC-P1), operating under a demanding set of joint constraints: sub-megabyte model footprint, sub-150\,ms per-batch TensorRT FP16 inference on an NVIDIA Jetson Xavier~NX, and an end-to-end alert pipeline targeting under 10 minutes from satellite overpass to fire event communication. The system operates on uncalibrated mid-wave infrared (MWIR) single-band imagery at 200\,m ground sampling distance, where fires frequently appear as sub-pixel or single-pixel thermal anomalies under extreme class imbalance---challenges not addressed by the contextual thermal-thresholding pipelines (MODIS, VIIRS) that currently dominate operational fire monitoring.

We present an empirical study of lightweight dense representation learning for this regime using a proprietary nine-satellite MWIR dataset. We compare dense masked autoencoding (DenseMAE) and a hybrid DenseMAE+EMA (exponential moving average) distillation variant, and evaluate representations via linear probing and full-distribution pixel-level average precision (AP) under extreme class imbalance. DenseMAE pretraining enables compact downstream models that sit on the latency--accuracy Pareto frontier: our fastest SSL-pretrained model achieves \textbf{0.640} test AP and \textbf{0.69} event-level Fire-F1 with \textbf{65.34\,ms} TensorRT FP16 latency per batch ($224\times224$, $B{=}8$) and a \textbf{0.52\,MB} engine, achieved without pruning or compression. The best-performing configuration reaches \textbf{0.699} test AP and \textbf{0.744} Fire-F1 while remaining below 1\,MB. Both outperform a supervised baseline (up to 0.650 AP) under comparable constraints.

\end{abstract}

\section{Introduction}
Wildfires are intensifying in frequency, severity, and seasonal duration, threatening ecosystems, infrastructure, and human life while contributing substantially to global CO$_2$ budgets \cite{IPCC_2021_WGI_Ch_11}. Rapid and reliable fire detection is therefore critical to enable early mitigation efforts and to support operational response systems.
Satellite-based surveillance has become a cornerstone of global fire monitoring, yet current architectures impose fundamental trade-offs. Geostationary platforms deliver high temporal revisit but at coarse spatial resolution (2–3~km), limiting sensitivity to small fires. Sensors onboard low Earth orbit (LEO) satellites achieve finer resolution but provide only a few daily overpasses, often missing the critical afternoon window of peak fire activity. OroraTech addresses this gap with a dedicated constellation of thermal infrared satellites (OTC-P1 mission). The constellation employs uncooled microbolometer sensors acquiring mid-wave infrared (MWIR) and long-wave infrared (LWIR) measurements at a ground sampling distance of 200\,m, combining the spatial sensitivity of LEO systems with substantially increased temporal coverage. With nine satellites in a late-afternoon sun-synchronous orbit, the constellation achieves approximately twice-daily revisit of fire-prone regions during the critical peak fire activity window. The entire pipeline is designed to deliver end-to-end alert latencies under 10 minutes from satellite overpass to fire event communication---a requirement that directly motivates on-board inference rather than downlink-and-process architectures.

This operational setting imposes compounding constraints. To meet real-time alert requirements, inference must execute entirely on board on an NVIDIA Jetson Xavier NX SoC (10\,W mode; up to 14 TOPS INT8, 384-core Volta GPU, 6-core Carmel CPU, 8\,GB LPDDR4x), making fire detection a high-stakes edge AI problem. To further minimise latency, the pipeline operates directly on uncalibrated MWIR measurements, and does not apply a radiometric calibration at the cost of increased susceptibility to sensor noise and radiometric artefacts. Compounding this, early-stage fires manifest as isolated sub-pixel or single-pixel thermal anomalies, producing extreme class imbalance and weak signals that are difficult to detect reliably with compact models trained on limited labelled data.
We address these challenges through a systematic study of lightweight dense representation learning, formulating on-orbit fire detection as dense segmentation on uncalibrated MWIR imagery under strict latency and resource constraints.
\paragraph{Contributions.}
Our main contributions are:
\begin{itemize}
    \item We introduce a deployment-oriented deep-learning pipeline for on-orbit wildfire detection from \emph{uncalibrated single-band} MWIR imagery under strict latency and footprint constraints.
    \item We propose \textbf{DenseMAE}, a lightweight staged convolutional masked autoencoder that learns dense MWIR representations and enables \textbf{sub-megabyte} deployment models after downstream fine-tuning.
    \item We provide a systematic empirical study of self-supervised pretraining (MAE vs. hybrid EMA distillation) and transfer (frozen vs. full fine-tuning) under extreme class imbalance, using \emph{pixel-level AP} computed on the full test distribution.
    \item We demonstrate that dense SSL representations enable
    \textbf{sub-megabyte deployment models} achieving competitive wildfire detection performance without pruning or compression.
\end{itemize}

\section{State of the Art}

Satellite-based fire detection exploits the characteristic increase in mid-wave infrared (MWIR, 3--4~$\mu$m) emission and MWIR--TIR (thermal infrared, 10--12~$\mu$m) contrast produced by combustion sources \cite{flasse_contextual_1996}. Operational algorithms therefore detect thermal anomalies relative to local background statistics, followed by rule-based filtering to suppress false positives such as sunglint or hot surfaces \cite{wooster_satellite_2021}. The MODIS Collection~6 product (MOD14/MYD14) formalized this contextual paradigm into a globally consistent workflow combining MWIR thresholds, background estimation, and sequential rejection tests \cite{giglio_collection_2016}. The VIIRS 375~m active-fire product inherits this logic while improving sensitivity to smaller fires due to higher spatial resolution \cite{schroeder_new_2014}. Similar contextual thresholding approaches are used in modern GEO imagers such as ABI \cite{2012_Schmidt_ABI}, or FCI~\cite{2026_Xu}, adapted to their coarser pixel sizes. Despite sensor improvements, these systems remain heuristic pipelines requiring sensor-specific tuning and hard decision rules.

Recent work explores learning-based alternatives that model spatial context directly from data. Classical machine-learning methods and CNN architectures have been applied to satellite imagery for fire detection \cite{2023_Ghalil,2022_Hong}. Encoder--decoder segmentation networks further enable pixel-level fire and smoke mapping \cite{2022_Seydi,2022_Wang}, while multi-scale or recurrent designs incorporate spatial and temporal context \cite{2022_Rostami,2022_Zhao}. However, existing approaches typically rely on calibrated multi-band inputs and comparatively large models, and therefore do not address the joint constraints of on-orbit inference, single-channel MWIR inputs, extreme class imbalance, and sub-pixel fire signatures considered in this work.

\section{Problem Setting and Dataset}
\label{sec:Dataset}

We train and evaluate on data acquired by the SAFIRE-2 instruments aboard FOREST-2 and the SAFIRE-3 aboard FOREST-4 through FOREST-11 of OroraTech's OTC-P1 mission~\cite{2026_book_cawse_nicholson}. Both instruments (SAFIRE-2/3) share the same dual-telescope uncooled micro-bolometer design, providing a 410\,km combined swath at 200\,m ground sampling distance across two LWIR channels and one MWIR channel (M0: centered at 3.8\,\textmu m, dynamic range extending to $\sim650$\,K). Detection relies exclusively on the uncalibrated M0 channel, expressed as raw digital numbers (DN) to minimise processing latency.

\begin{figure}[tb]
  \centering
  \includegraphics[width=\linewidth]{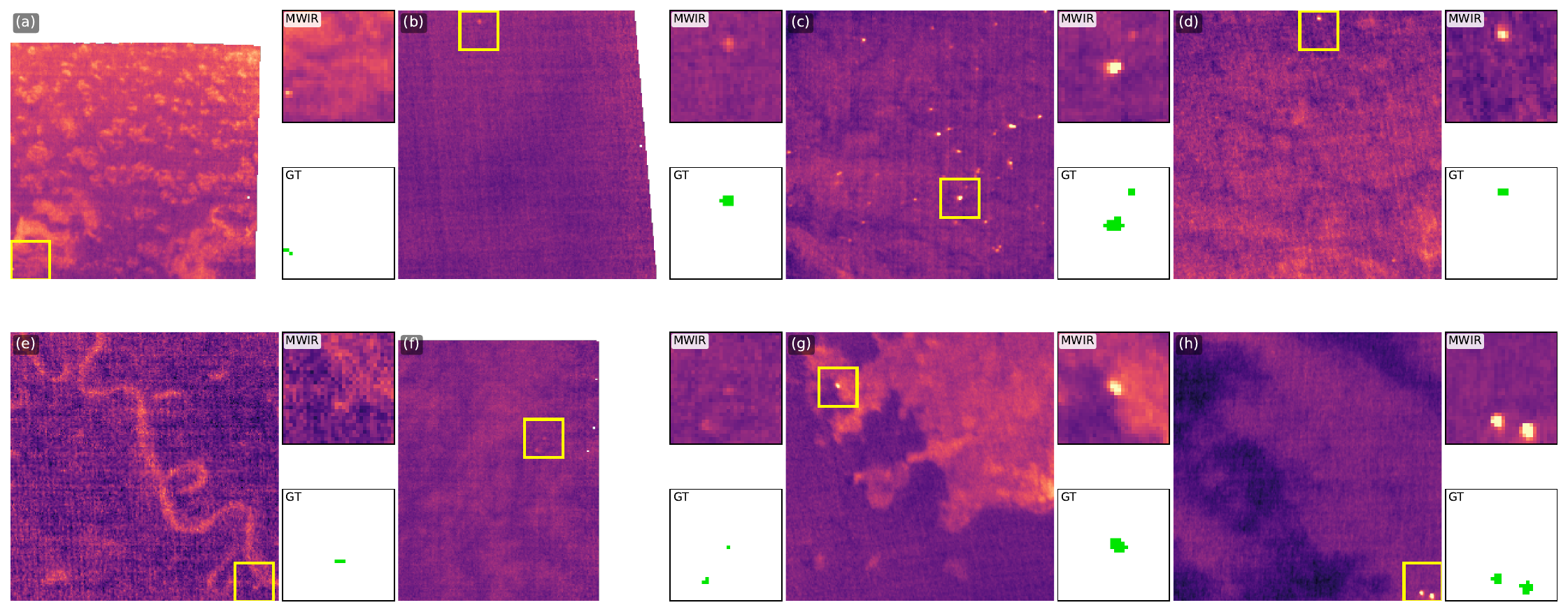}
  \caption{Random sample tiles of the MWIR band along with the manually annotated ground truth.}
  \label{fig:refdataset_samples81}
\end{figure}

The dataset contains 839 annotated scenes (2023-08-01 to 2025-08-18), each with at least one confirmed fire. Scenes are split into 589/126/124 train/val/test recordings (70/15/15\%); we use 5-fold cross-validation within the training split during development while keeping the test split fixed for final reporting. Scenes are tiled into $224\times224$ patches (16-pixel overlap). The data is extremely imbalanced: fire pixels constitute 0.018\% of valid training pixels and only 10.86\% of training patches contain any fire; 19--21\% of pixels are invalid and ignored (Table~\ref{tab:dataset_split_stats}). Pixel values are normalised per scene using a robust scaler over valid pixels, and we apply lightweight spatial augmentations (flips and $90^\circ$ rotations) during training. Figure~\ref{fig:refdataset_samples81} illustrates example tiles, highlighting that many fires occupy only a few pixels in noisy backgrounds.

\begin{table}[t]
\centering
\caption{Split-wise class imbalance statistics after tiling ($224\times224$ patches, overlap $16\times16$). Positive pixel ratio is computed over valid pixels only. "Ignored Pixels" could be masked sensor artefacts. These pixels will be ignored (treated as no-data values)}
\label{tab:dataset_split_stats}
\small
\setlength{\tabcolsep}{5pt}
\begin{tabular}{lrrrr}
\toprule
Split & \# Patches & Positive Patches (\%) & Positive Pixels (\%) & Ignored Pixels (\%) \\
\midrule
Train & 49{,}141 & 10.86 & 0.0182 & 21.30 \\
Val   & 10{,}634 &  7.34 & 0.0058 & 19.40 \\
Test  &  9{,}592 &  9.37 & 0.0112 & 21.16 \\
\bottomrule
\end{tabular}
\end{table}

\section{Methodology}
\label{sec:Methodology}

We describe the model architectures and learning objectives used in our study. The problem setting, dataset, and evaluation splits are defined in Section~\ref{sec:Dataset}; here we focus on (i) the supervised U-Net++ baseline, (ii) DenseMAE self-supervised pretraining, (iii) the hybrid distillation ablation, and (iv) transfer to downstream segmentation with lightweight heads.

\subsection{Supervised Reference Model for Downstream Segmentation}
As a production-oriented reference, we use a convolutional encoder--decoder segmentation model based on U-Net++~\cite{zhou2018unetpp} with an EfficientNet backbone~\cite{tan2019efficientnet}. U-Net++ is a strong dense-prediction baseline due to its nested skip connections and multi-scale feature fusion.
We selected this family after an internal architecture sweep over common encoder--decoder variants, where U-Net++ offered the best accuracy--latency trade-off under our deployment constraints (details in the supplementary material).
To improve sensitivity to tiny fire signatures, we extend the base model with a lightweight high-resolution refinement head.

Let $f_0$ denote the highest-resolution encoder feature map and $L_c$ the coarse decoder logits. The refinement head predicts final logits as
\[
L_r = \phi([f_0, L_c]),
\]
where $\phi$ is a shallow convolutional module and $[\cdot,\cdot]$ denotes channel-wise concatenation. This design re-injects high-resolution encoder detail into the final prediction stage and improves localization of single-pixel or sub-pixel fire responses. We optionally apply deep supervision by supervising downsampled variants of the refined logits. In practice, the supervised reference model is trained with binary cross-entropy on valid pixels, and serves as the main downstream baseline for comparison with embedding-based approaches.

\paragraph{Why we compare to a strong internal baseline.}
Direct comparison to published deep-learning fire segmentation models is challenging in our setting because most prior work assumes calibrated multi-spectral inputs (often including visible/NIR bands) and does not target uncalibrated single-band MWIR with on-orbit compute constraints. We therefore compare against (i) a strong production-oriented U-Net++ baseline selected via an internal architecture sweep, and (ii) external VIIRS active-fire detections as a widely used operational reference (Section~\ref{sec:results_speed} and Section~\ref{sec:ResultsandDiscussion}).

\subsection{DenseMAE for MWIR Embedding Pretraining}

To study data-efficient representation learning for this domain, we pretrain a dense masked autoencoder (DenseMAE) tailored to MWIR imagery and dense downstream tasks. Unlike standard MAE variants that operate on tokenized image patches, DenseMAE is fully convolutional and employs a staged encoder with a single early downsampling step for computational efficiency.

\paragraph{Architecture.}
\begin{figure}
    \centering
    \includegraphics[width=1\linewidth,trim=0 220 0 40, clip]{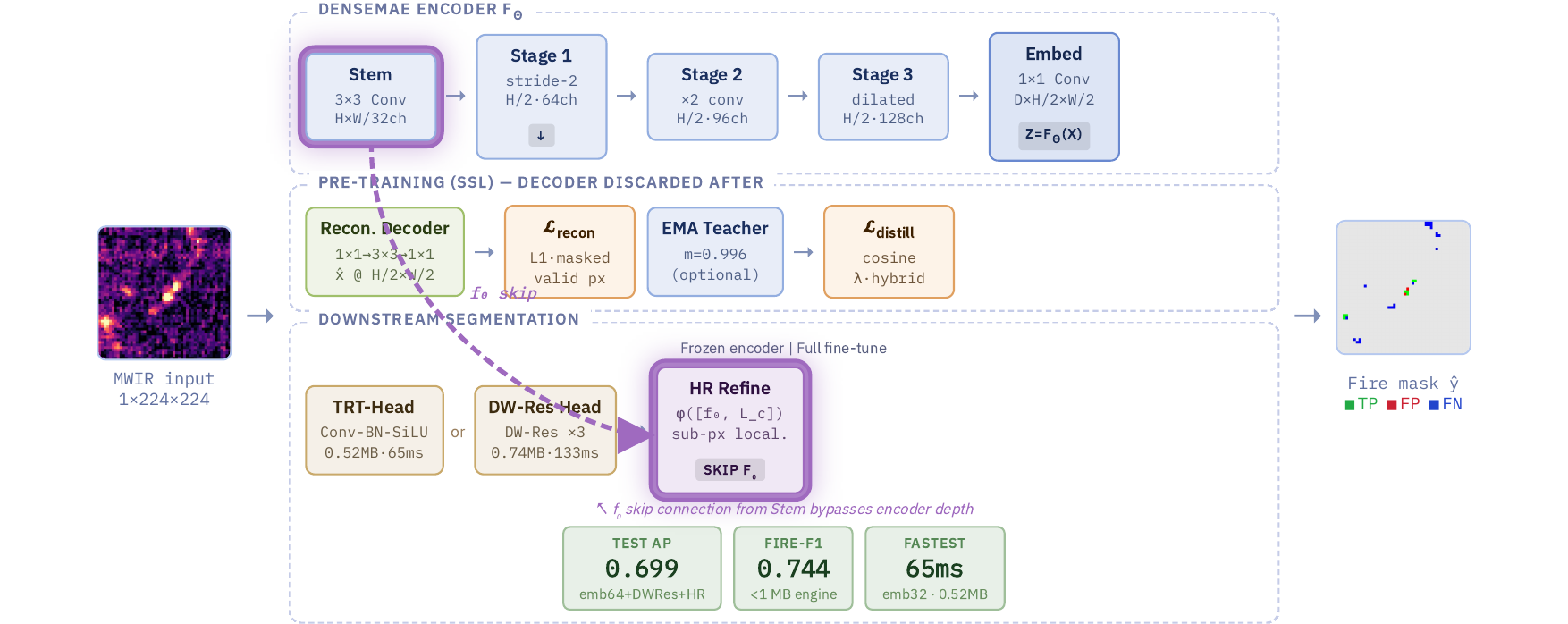}
    \caption{\textbf{DenseMAE pipeline.} Staged convolutional encoder pre-trained with masked autoencoding on uncalibrated MWIR imagery, transferred to wildfire segmentation via a lightweight head and HR refinement. The f$_0$ skip connection (purple) re-injects high-resolution stem features for sub-pixel fire localisation.}
    \label{fig:placeholder}
\end{figure}
While modern segmentation systems can be substantially larger, our focus is on the efficient regime relevant to edge/on-board deployment, where lightweight designs are actively studied \cite{2022_Holder}. DenseMAE is a lightweight fully convolutional masked autoencoder designed to produce dense feature maps for downstream segmentation. It consists of (i) a high-resolution $3\times3$ stem, (ii) a staged encoder that downsamples once to $H/2\times W/2$ and then applies dilated blocks, and (iii) a lightweight reconstruction decoder used only during pretraining. Given an input tile $x \in \mathbb{R}^{1 \times H \times W}$, the encoder produces a dense embedding map
\[
z = f_\theta(x) \in \mathbb{R}^{D \times \frac{H}{2} \times \frac{W}{2}},
\]
where $D$ is the embedding dimensionality (we ablate $D\in\{32,64,128\}$). After the initial downsampling stage, the encoder uses dilated blocks with dilation rates $\{1,2,4\}$ to increase receptive field while operating on the reduced grid; fine detail is recovered downstream by fusing high-resolution features via the refinement head (Section~\ref{sec:Methodology}, Supervised Reference Model). Throughout the encoder we use GroupNorm (8 groups) and GELU activations for stability under small batch sizes. Table~\ref{tab:densemae_spec} provides the detailed staged encoder specification used in our implementation.

\begin{table}[t]
\centering
\caption{DenseMAE architecture specification (matching our implementation). The staged encoder downsamples once to $H/2\times W/2$ for efficiency. A \emph{conv unit} denotes $3\times3$ Conv + Norm + Activation.}
\label{tab:densemae_spec}
\small
\begin{tabular}{@{}lll@{}}
\toprule
\textbf{Stage} & \textbf{Operation} & \textbf{Resolution / Out ch.} \\
\midrule
Stem            & conv unit ($3\times3$) & $H\times W$ / 32 \\
Stage 1         & $3\times3$ Conv (stride=2) + conv unit (d=1) & $H/2\times W/2$ / 64 \\
Stage 2         & conv unit (d=1) $\times2$ & $H/2\times W/2$ / 96 \\
Stage 3         & conv unit (d=2) $\times1$ + conv unit (d=1) $\times1$ & $H/2\times W/2$ / 128 \\
Embedding proj. & $1\times1$ Conv + Activation & $H/2\times W/2$ / $D$ \\
Decoder (pretrain) & $1\times1 \to 3\times3 \to 1\times1$ Conv & $H/2\times W/2$ / $64 \to 64 \to 1$ \\
\bottomrule
\end{tabular}
\end{table}

The reconstruction decoder maps $z$ to a reconstruction $\hat{x}$ and is intentionally lightweight. After pretraining, we discard the decoder and reuse only the encoder to provide dense per-pixel features for downstream segmentation.
\paragraph{Masking and reconstruction objective.}
During pretraining, we apply random block masking in image space with mask ratio $r$ and block size $b$ (ablated in Section~\ref{sec:ResultsandDiscussion}). Let $M \in \{0,1\}^{1\times H\times W}$ denote the binary mask (1 = masked), and let $V$ denote the valid-pixel mask derived from the no-data map. Since the reconstruction is predicted at $H/2\times W/2$, we downsample the input $x$ to $x_{\downarrow2}$ using area interpolation, and downsample $M$ and $V$ with nearest-neighbor interpolation before computing the loss. The reconstruction loss is computed only on masked, valid pixels:
\[
\mathcal{L}_{\text{recon}} =
\frac{\lVert (\hat{x} - x_{\downarrow2}) \odot M \odot V \rVert_1}{\sum (M \odot V) + \epsilon}.
\]
This objective encourages the encoder to infer missing thermal patterns from surrounding context while avoiding no-data regions.

\paragraph{Differences of DenseMAE to standard MAE.}
The proposed DenseMAE is designed for dense remote-sensing transfer rather than image-level classification. In particular, it (i) produces dense embeddings on a regular grid with a single early downsampling step for efficiency, (ii) uses dilated convolutions instead of patch-token self-attention, (iii) explicitly excludes invalid pixels from the objective, and (iv) produces dense per-pixel embeddings that can be directly fine-tuned for downstream segmentation.

\subsection{Hybrid SSL Extension: MAE with EMA Distillation}
\label{sec:hybrid_ablation}

\emph{This section describes a diagnostic ablation, not a core contribution.} We include it to explicitly test whether adding a consistency objective improves representation quality beyond reconstruction-only DenseMAE in our noise-dominated MWIR regime. The results (Section~\ref{sec:results_ssl}) show only marginal and inconsistent gains, confirming that DenseMAE reconstruction alone provides the stronger foundation.

In addition to MAE-only pretraining, we evaluate a hybrid self-supervised variant that combines masked reconstruction with a teacher--student consistency objective. The student is a DenseMAE model trained with reconstruction loss, while the teacher is an exponential moving average (EMA) copy of the student encoder.

Let $z_s$ and $z_t$ denote student and teacher embeddings, respectively. After channel-wise normalization, we minimize a cosine consistency loss on valid pixels:
\[
\mathcal{L}_{\text{distill}} =
\frac{\sum \left(1 - \cos(\theta(z_s, z_t))\right) \odot W}{\sum W + \epsilon},
\]
where $W$ is either the masked region only or all valid pixels (ablation) and $\theta(x, y)$ is the angle between vectors $x$ and $y$. The total SSL loss is
\[
\mathcal{L}_{\text{SSL}} = \mathcal{L}_{\text{recon}} + \lambda\,\mathcal{L}_{\text{distill}},
\]
with distillation weight $\lambda$ and EMA momentum $m$ treated as hyperparameters.

This hybrid formulation is included as an ablation to test whether consistency regularization improves MWIR transfer performance beyond reconstruction-only DenseMAE pretraining.

\subsection{Using Pretrained Embeddings for Downstream Fire Segmentation}

After SSL pretraining, we discard the reconstruction decoder and transfer the DenseMAE encoder to the downstream wildfire segmentation task. We evaluate multiple lightweight segmentation heads on top of the dense embedding map $z$ to isolate the effect of representation quality from decoder capacity.

\paragraph{Embedding-based heads.}
We consider the following downstream heads:
\begin{itemize}
    \item \textbf{Linear head:} a single $1\times1$ convolution (linear probe baseline).
    \item \textbf{Context head:} a small convolutional head with local context aggregation (e.g., $1\times1 \rightarrow 3\times3 \rightarrow 1\times1$) and normalization/dropout.
    \item \textbf{Depthwise-residual variants:} lightweight spatial heads with depthwise separable convolutions for improved efficiency.
\end{itemize}

We evaluate two transfer settings that reflect different operational trade-offs. 
\begin{description}
    \item[(i) Frozen-encoder transfer] We freeze the pretrained DenseMAE encoder and train only a lightweight head on top of $z$ (linear probe and context heads). This isolates representation quality and enables fast adaptation with minimal compute costs.
    \item[(ii) Full fine-tuning] We jointly optimize encoder and head end-to-end on the downstream segmentation objective. This setting tests whether SSL initialization improves final task performance compared to training the same architecture from scratch.
\end{description} 

\paragraph{Lightweight segmentation heads}
To isolate representation quality from decoder capacity, we attach compact heads on top of DenseMAE embeddings. (i) TRT-Head is an ultra-light head composed only of Conv–BatchNorm–SiLU blocks (1×1 projection, a small number of 3×3 Conv–BN–SiLU blocks, and a 1×1 output), designed to maximize operator fusion and throughput in TensorRT. (ii) DW-Res Context Head increases representational capacity while remaining lightweight by projecting embeddings to a hidden width and applying a small stack of depthwise-separable residual blocks with dilated depthwise convolutions, followed by a 1×1 output layer. We optionally apply per-pixel L2 normalization of embeddings before the head for stability. 

\subsection{Training Strategy and Imbalance Handling}
Wildfire pixels represent only a minute fraction of all valid pixels (Table~\ref{tab:dataset_split_stats}), making naive optimisation unstable. We therefore bias training batches toward positive patches, exclude no-data pixels from all losses and metrics, and optimise with binary cross-entropy using AdamW and mixed precision. For SSL pretraining, we ablate embedding dimensionality, mask ratio, and mask block size; for hybrid SSL, we additionally ablate EMA momentum and distillation weight.
\subsection{Evaluation Protocol}
\label{sec:eval_protocol}
We report pixel-level average precision (AP) computed over all valid pixels in the validation/test splits, reflecting operational prevalence (Table~\ref{tab:dataset_split_stats}). For SSL checkpoint selection we use a fixed \emph{probe} protocol: we sample a static pool of tiles per split once (fixed seed/indices), extract up to $N_{+}$ positive pixels and $kN_{+}$ negatives from valid background, train a linear classifier on the training pool, and report AP on the validation pool. Probe AP is used only to rank checkpoints; all headline numbers are full-stream test AP and event-level Fire-F1.

\paragraph{Probe AP (checkpoint selection).}
To rank SSL checkpoints efficiently, we compute probe AP using a fixed embedding pool sampled once per split and reused for all checkpoints. Concretely, we sample $N_{\text{tiles}}$ tiles from the training split and extract up to $N_{+}$ positive pixels and $kN_{+}$ negative pixels (uniform over valid background) to form a training pool; we repeat the same procedure on the validation split to form a validation pool. Sampling uses a fixed random seed and fixed tile indices to ensure comparability across checkpoints. A linear classifier is trained on the training pool and evaluated on the validation pool to obtain \emph{probe AP}. We use probe AP only for checkpoint ranking; all headline results are based on full-stream pixel AP and event-level Fire-F1 on the fixed test split.

\paragraph{Event-level metric (Fire-F1).}
Pixel AP can be dominated by large fire clusters. We therefore report an event-level score using connected components (CCs) on the ground truth (8-connectivity, valid pixels only). Each CC is treated as one fire event. Given a thresholded prediction at $t$, an event is counted as detected if any predicted positive pixel intersects its CC. We compute event-level recall as the fraction of ground-truth events detected. For precision, we compute CCs in the prediction and count a predicted CC as a true positive if it overlaps any ground-truth CC; otherwise it is a false positive. Fire-F1 is the harmonic mean of event precision and recall. The threshold $t$ is chosen on the validation split by maximizing validation Fire-F1 and then fixed for test reporting.

\section{Results and Discussion}
\label{sec:ResultsandDiscussion}

We evaluate our wildfire detection pipeline under constraints motivated by on-orbit deployment: ultra-low latency alerting, strict model footprint limits, and extreme class imbalance. Unless stated otherwise, average precision (AP) is computed on the \emph{full} validation/test distributions (all valid pixels), reflecting the operational prevalence reported in Table~\ref{tab:dataset_split_stats}. We structure results to (i) analyze representation learning choices, (ii) compare downstream segmentation performance, and (iii) quantify deployment trade-offs (latency vs.\ accuracy). Our experiments show that DenseMAE representations transfer effectively to downstream wildfire segmentation and outperform the supervised baseline under comparable deployment constraints. The best configuration achieves 0.69 AP while remaining within the latency and memory budgets required for on-orbit inference.
\begin{figure}
    \centering
    \includegraphics[width=1\linewidth]{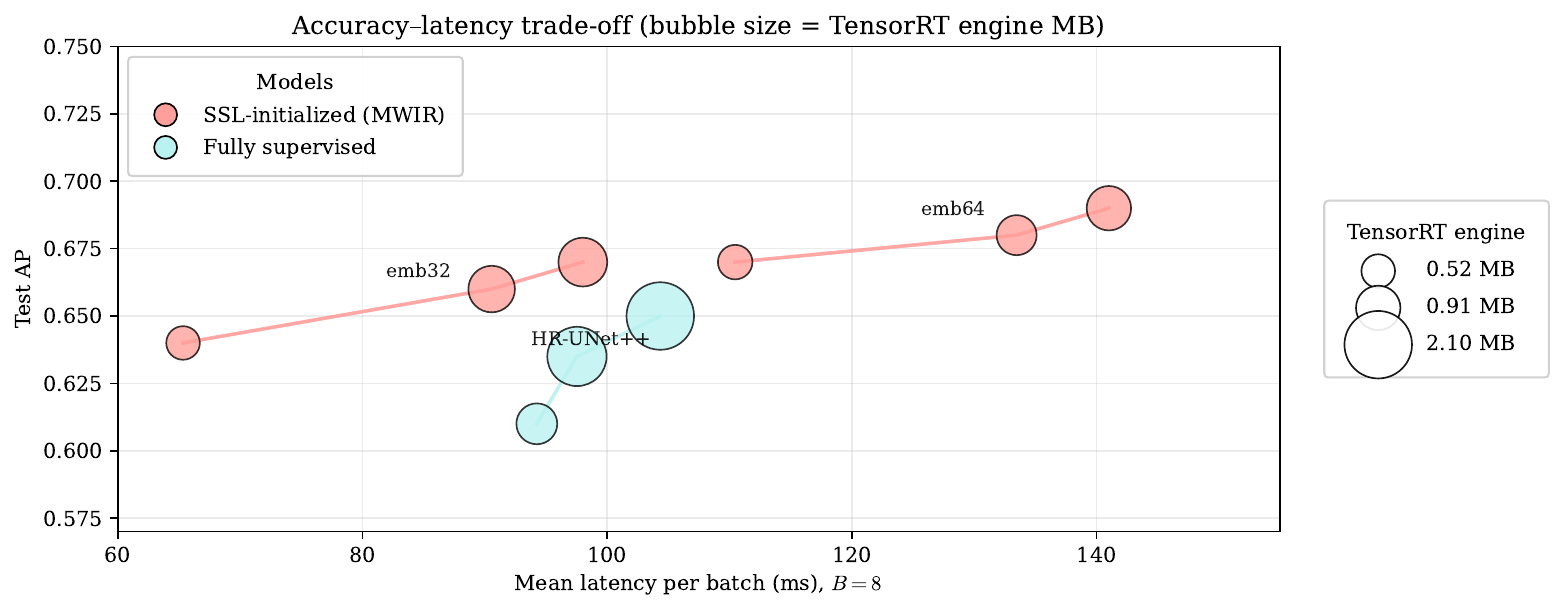}
\caption{Latency--accuracy trade-off for deployment models (TensorRT FP16, $224\times224$, $B{=}8$). Points on the upper-left boundary are Pareto-optimal.}
    \label{fig:pareto}
\end{figure}
\subsection{Self-supervised pretraining study: DenseMAE vs.\ Hybrid}
\label{sec:results_ssl}

We first study how different self-supervised objectives affect representation quality on MWIR imagery. We evaluate DenseMAE (masked reconstruction) and a hybrid variant that adds EMA self-distillation (Section~\ref{sec:Methodology}). We report two complementary indicators: (i) a \emph{probe} AP measured on a sampled pool of embeddings for rapid monitoring, and (ii) \emph{full-stream} AP measured over the full validation/test splits using a lightweight downstream head.

\begin{table*}[t]
\centering
\caption{Self-supervised pretraining ablations (test split). We select checkpoints by \textbf{validation probe AP} (fast proxy on a sampled pool), then evaluate \textbf{full-stream AP} over all valid pixels at operational prevalence.}
\label{tab:ssl_ablation}
\small
\setlength{\tabcolsep}{5pt}
\begin{tabular}{llcccccc}
\toprule
\textbf{Method} & \textbf{Variant} &
$D$ & $r$ & $b$ &
$\lambda$ & $m$  &
\textbf{Test AP} $\uparrow$ \\
\midrule
MAE & MAE & 32 & 0.60 & 4 & --  & -- & 0.1137 \\
MAE & MAE & 64 & 0.75 & 4 & --  & -- & 0.1502 \\
MAE & MAE & 128 & 0.60 & 4 & -- & -- & 0.1593 \\
\addlinespace
MAE & MAE & 64 & 0.60 & 2 & --   & --  & \textbf{0.1812} \\
MAE & MAE & 64 & 0.60 & 4 & --   & --  & 0.1612\\
MAE & MAE & 64 & 0.60 & 8 & --   & --  & 0.1251 \\
\midrule
Hybrid & hybrid & 32 & 0.60 & 2 & 0.02 & 0.996 &  0.2655 \\
Hybrid & hybrid & 64 & 0.60 & 2 & 0.02 & 0.996 & 0.2796 \\
Hybrid & hybrid & 128 & 0.60 & 2 & 0.02 & 0.996 & \textbf{0.2869}  \\
\addlinespace
Hybrid & hybrid & 32 & 0.60 & 4 & 0.02 & 0.996 & 0.255 \\
Hybrid & hybrid & 32 & 0.60 & 4 & 0.5 & 0.996  & 0.1828 \\
\bottomrule
\end{tabular}
\end{table*}

Overall, DenseMAE representations are stable across ablations and consistently transfer to downstream segmentation. Hybrid pretraining provides only marginal improvements in the best case, and we observe that applying distillation on masked regions only can degrade \textbf{full-stream} performance (Table~\ref{tab:ssl_ablation}), suggesting that reconstruction already provides a strong learning signal for this noise-dominated MWIR regime.

\subsection{Downstream wildfire segmentation: supervised vs.\ representation learning}
\label{sec:results_downstream}

We compare (i) a production-oriented supervised baseline (U-Net++ + high-resolution refinement; Section~\ref{sec:Methodology}) against (ii) DenseMAE-based segmentation where the pretrained encoder produces dense embeddings and a lightweight head predicts the fire mask. We evaluate two transfer settings: \emph{frozen-encoder transfer} (train head only) and \emph{full fine-tuning} (encoder + head). Table~\ref{tab:downstream_main} summarises the main downstream segmentation results, contrasting supervised training against DenseMAE initialisation under frozen-encoder and full fine-tuning transfer.

\begin{table}[t]
\centering
\caption{Downstream segmentation performance under extreme class imbalance.
Average precision (AP) is computed on the full test distribution (all valid pixels).
We report mean $\pm$ standard deviation across five cross-validation folds.}
\label{tab:downstream_main}
\small
\setlength{\tabcolsep}{4pt}
\begin{tabular}{lcc}
\toprule
\textbf{Model} & \textbf{Transfer} & \textbf{Test AP} $\uparrow$ \\
\midrule
U-Net++ + HR refine (supervised) & end-to-end & $0.650 \pm 0.006$ \\
DenseMAE (emb32) + lightweight head & fine-tune & $0.671 \pm 0.005$ \\
DenseMAE (emb64) + lightweight head & fine-tune & $\mathbf{0.689 \pm 0.004}$ \\
DenseMAE (emb64) + lightweight head & frozen encoder & $0.612 \pm 0.006$ \\
DenseMAE (emb64) + random init & fine-tune & $0.551 \pm 0.008$ \\
\bottomrule
\end{tabular}
\end{table}

DenseMAE initialization yields modest improvements over a supervised baseline
while maintaining compact model sizes suitable for deployment. Increasing the
embedding dimensionality from 32 to 64 consistently improves AP, indicating
higher representational capacity in the encoder.

\begin{figure}[t]
    \centering
    \includegraphics[width=1\linewidth]{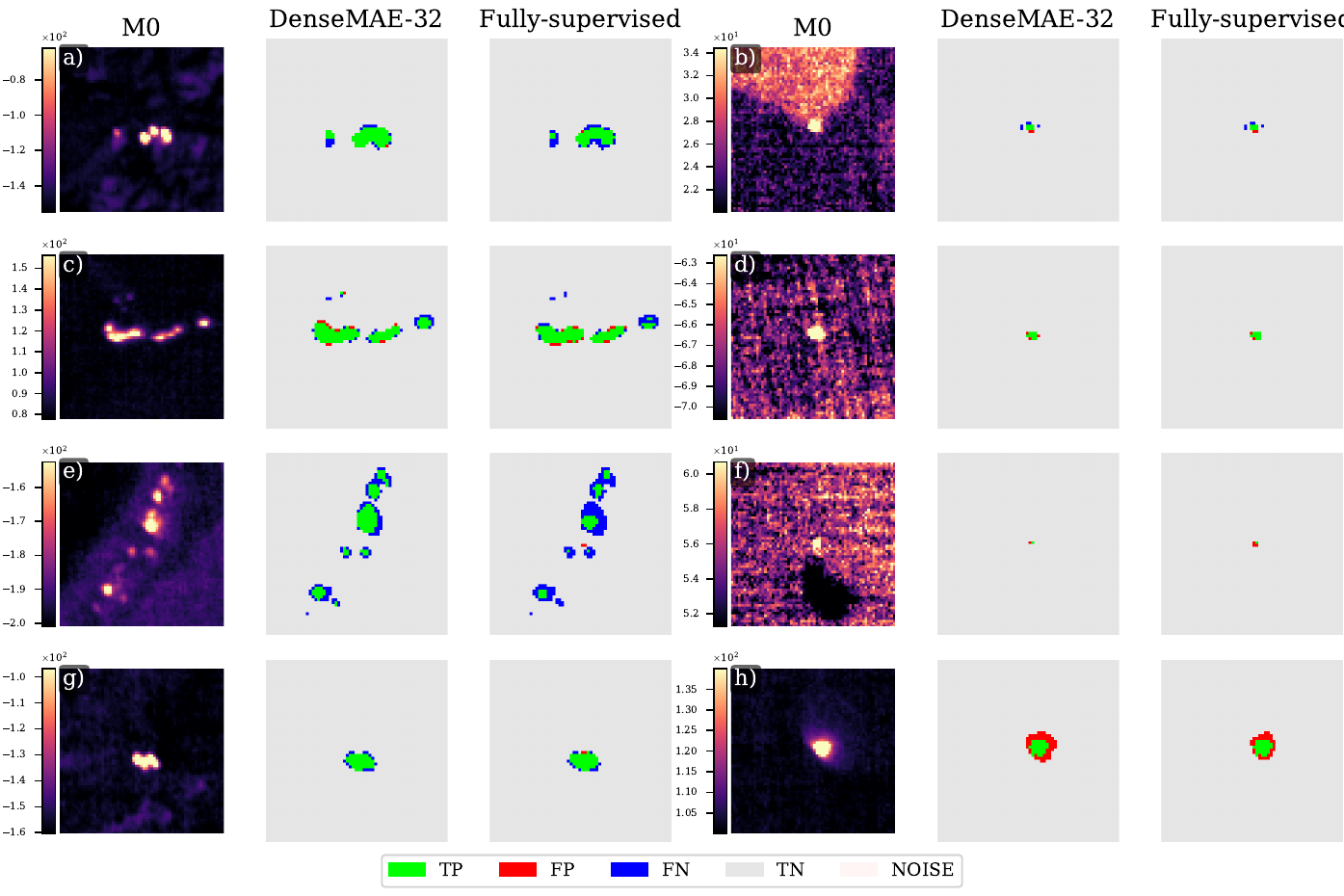}
    \caption{Qualitative comparison on challenging MWIR tiles ($51 \times 51$ crops).}
    \label{fig:qualitative}
\end{figure}

To complement AP under extreme imbalance, we include a qualitative comparison focusing on small and isolated fire signatures (Fig.~\ref{fig:qualitative}). We highlight cases where dense representations improve sensitivity to weak MWIR anomalies while suppressing common false positives from structured sensor artefacts.

\subsection{Deployment feasibility: latency--accuracy trade-offs}
\label{sec:results_speed}

On-orbit feasibility is constrained by inference latency and model footprint (engine size). We report TensorRT FP16 latency on $224\times224$ tiles at batch size $B=8$ as a proxy for deployment performance. Table~\ref{tab:pareto_latency_ap} summarizes the latency--accuracy frontier and Fig.~\ref{fig:pareto} visualizes the Pareto trade-off.

All latency measurements are obtained from TensorRT engines compiled in FP16.
We compare FP16 TensorRT outputs against an ONNX Runtime FP32 reference on a
shared set of validation tiles and observe negligible deviations in downstream
metrics ($\Delta$AP $\approx 0.5$ percentage points).

\begin{table}[t]
\centering
\caption{Latency--accuracy trade-off across fully supervised baselines and DenseMAE-based embedding models. Latency is measured using TensorRT FP16 engines ($224\times224$, batch $B=8$).We additionally report event-level Fire-F1 based on connected components, using a decision threshold selected on the validation split and then fixed for test reporting (see Section~\ref{sec:eval_protocol}).}
\label{tab:pareto_latency_ap}
\small
\begin{tabular}{lcccc}
\toprule
\textbf{Model} & \textbf{Latency (ms)} $\downarrow$ & \textbf{AP} $\uparrow$ & \textbf{Fire-F1} $\uparrow$ & \textbf{TRT Eng. (MB)} $\downarrow$ \\
\midrule
\multicolumn{5}{l}{\emph{Fully supervised baselines}} \\
HR-U-Net++ (depth=2, h=32) & 94.25 & 0.61 & 0.68 & 0.77 \\
HR-U-Net++ (depth=3, h=32) & 97.53 & 0.635 & 0.69 & 1.62 \\
HR-U-Net++ (depth=4, h=32) & 104.34 & 0.65 & 0.73 & 2.10 \\
\midrule
\multicolumn{5}{l}{\emph{DenseMAE embeddings (emb\_ch=32)}} \\
TRT-Head & \textbf{65.34} & 0.640 & 0.69 & \textbf{0.52} \\
DW-Res-Head & 90.56 & 0.661 & 0.701 & 1.00 \\
DW-Res-Head + hr-refine & 98.01 & 0.677 & 0.732 & 1.10 \\
\midrule
\multicolumn{5}{l}{\emph{DenseMAE embeddings (emb\_ch=64)}} \\
TRT-Head & 110.46 & 0.677 & 0.689 & 0.55 \\
DW-Res-Head & 133.46 & 0.689 & 0.711 & 0.74 \\
DW-Res-Head + hr-refine & 141.00 & \textbf{0.699} & \textbf{0.744} & 0.91 \\
\bottomrule
\end{tabular}
\end{table}
\vspace{1mm}
\footnotesize
\emph{Head configuration:} TRT-Head: $1\times1$ proj $\rightarrow$ 3$\times$(3$\times$3 Conv-BN-SiLU) $\rightarrow$ $1\times1$ out with hidden width $32$.
DW-Res Head: $1\times1$ proj $\rightarrow$ 3$\times$(DW-Res block) $\rightarrow$ $1\times1$ out with hidden width $32$.

The results reveal a clear latency--accuracy trade-off between fully supervised and embedding-based variants. The fastest DenseMAE configuration ($c_{emb}=32$ with the fast head) achieves substantially lower latency than the supervised U-Net++ baseline while maintaining competitive AP. Increasing the embedding dimensionality from 32 to 64 improves AP (up to 0.69 with DWRes + HR refine), at the cost of higher latency. These experiments demonstrate that dense self-supervised pretraining enables compact models that sit on the Pareto frontier for on-orbit fire detection.
\paragraph{Why refinement matters for small fires.}
While pixel-level AP is dominated by large fire clusters, the high-resolution refinement head is designed to improve localisation of small and isolated fire signatures that contribute few pixels. Consistent with this goal, we observe larger gains in event-level Fire-F1 (Table~\ref{tab:pareto_latency_ap}) than in pixel-level AP, indicating that refinement particularly improves detection of small fire events.

\paragraph{Limitations.}
A persistent failure mode is confusion with sunglint and other reflective artefacts, which can produce MWIR responses similar to weak fires when operating on uncalibrated single-band inputs. While DenseMAE pretraining improves robustness to common sensor artefacts, rare edge cases remain challenging. Addressing these cases likely requires incorporating additional spectral channels (e.g., LWIR) or physics-informed features (illumination/viewing geometry), which we leave for future work. Temporal context is not exploited: the single-overpass constraint means that each scene is processed independently without access to prior frames, and the memory and inter-process synchronization budget of the Jetson Xavier NX precludes maintaining a persistent frame buffer across downlink cycles. Extending to multi-temporal inputs would require architectural changes and a relaxation of the current memory footprint constraints.

\subsection{External Validation against VIIRS}
\label{sec:viirs_validation}
VIIRS represents the closest available operational baseline for our system. Direct numerical comparison with traditional contextual-thresholding algorithms is methodologically non-trivial in our setting: those algorithms rely on calibrated multi-spectral inputs (MWIR brightness temperatures, MWIR--TIR contrast), whereas our model operates on uncalibrated single-band DNs without radiometric correction. We therefore treat the VIIRS 375\,m active-fire product \cite{giglio2024viirs_vj114img}---a globally validated, extensively benchmarked reference~\cite{2020_VIIRS_I-Band_algorithm, 2014_SUOMI_NPP_Csiszar, 2015_VIIRS_assessment_OLIVA2015144, 2015_VIIRS_campaign_Dickinson} widely used in satellite fire monitoring~\cite{2026_Himawari_VIIRS_Liu, 2020_GOES_VIIRS_Li}---as an independent operational reference to assess cross-sensor agreement and real-world generalisation.

Coincident observations are defined by a $\pm$10\,min temporal window. To account for geolocation uncertainty and the resolution difference between our system (200\,m) and VIIRS (375\,m), detections are buffered by 800\,m and overlapping buffers merged into fire clusters, enabling event-level rather than pixel-level comparison.

\begin{figure}[tb]
  \centering
  \includegraphics[width=\linewidth]{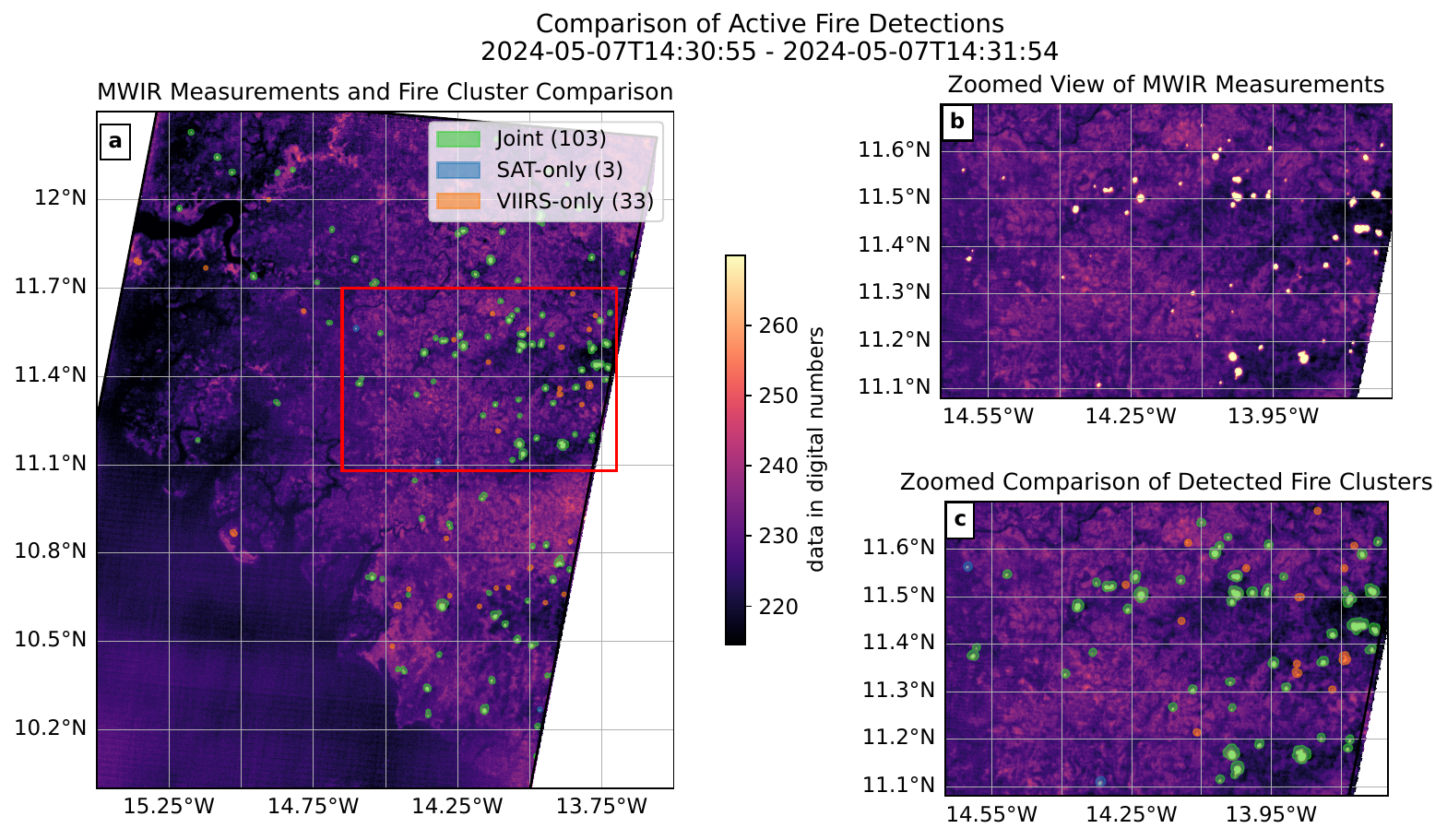}
  \caption{Comparison of active fire detections from our model and VIIRS. Panel a: MWIR measurements (in DNs) with evaluation of joint detections (green), model-only (blue), and VIIRS-only (orange). Panel b: Zoomed view of the MWIR measurements (indicated by the red rectangle). Panel c: Zoomed clustered detections.}
  \label{fig:viirs_example}
\end{figure}

Figure~\ref{fig:viirs_example} shows a representative scene. Panel (a) shows MWIR measurements (digital numbers) used as input to the model. As the model operates directly on uncalibrated data, sensor artefacts such as across-track striping (visible near 10.5°N) are present in the input space. In addition, the cluster-level comparison is shown: 103 joint detections (green), 33 VIIRS-only clusters (orange), and 3 model-only clusters (blue). The zoomed panels (b--c) indicated by the red rectangle in panel a reveal weak thermal signals at several VIIRS-only locations, suggesting fires near the model detection threshold, while model-only clusters coincide with clear thermal anomalies, indicating potentially small or short-lived fires below the VIIRS detection limit. Discrepancies between the systems can arise from differences in spatial resolution, viewing geometry, spectral response, and acquisition timing. Given the fundamental input mismatch---calibrated multi-spectral VIIRS versus uncalibrated single-band DNs in our system---this comparison is best interpreted as a cross-sensor agreement and generalisation assessment using the strongest operational reference available, rather than a pixel-accurate benchmark.

\section{Outlook}
While our results demonstrate the feasibility of lightweight deep learning for on-orbit wildfire detection, several directions remain for improving performance and robustness. Using calibrated radiances instead of raw digital numbers may reduce noise and improve sensitivity to weak thermal anomalies, though the additional latency must be balanced against detection gains. Expanding the dataset to underrepresented satellites, regions, and seasons will further improve generalisation. Incorporating multi-channel inputs and physically informed features such as illumination and viewing geometry may help distinguish fires from confounders such as sunglint. Finally, hybrid pipelines combining low-latency onboard detection with more computationally intensive ground-based refinement offer a promising path to further reduce end-to-end alert latency.

\section{Conclusion}

In this study we investigated lightweight dense representation learning for on-orbit wildfire detection, tailored to OroraTech's OTC-P1 thermal satellite constellation, which operates under a rare confluence of challenges: ultra-low latency requirements, severely resource-constrained hardware, sub-pixel fire  signatures, and extreme class imbalance in uncalibrated MWIR imagery.

DenseMAE pretraining yields strong, transferable representations for this regime. 
The best configuration (embedding dimension 64, lightweight head, high-resolution 
refinement) achieves 0.699 AP on the full test distribution, outperforming a 
production supervised U-Net++ baseline (0.650 AP) under comparable constraints. 
Full fine-tuning consistently outperforms frozen-encoder transfer (0.610 AP), 
indicating that DenseMAE initialisation is most beneficial when the downstream 
model can adapt end-to-end. Across ablations of embedding size, masking ratio, 
and block size, reconstruction-based pretraining is stable, while hybrid EMA 
distillation does not provide consistent full-distribution gains, suggesting that 
masked reconstruction already constitutes an effective learning signal in this 
noise-dominated setting.

DenseMAE models occupy a favourable deployment frontier: a compact configuration 
achieves 0.640 AP with 65.34\,ms TensorRT FP16 latency per batch ($B{=}8$) and a 
0.52\,MB engine, substantially reducing inference time and memory footprint 
relative to supervised baselines. External validation against coincident VIIRS 
active-fire products further demonstrates strong cross-sensor agreement, 
supporting generalisation beyond the annotated reference dataset.

With wildfire seasons intensifying globally, our results demonstrate that 
light\-weight self-supervised representations enable accurate and efficient 
on-orbit fire detection. Being able to detect ignitions within minutes 
from orbit represents a significant step toward reducing the gap between fire 
onset and operational response.

\section*{Acknowledgements}
We acknowledge the work of the Optical Mission Performance Cluster (OPT-MPC) done within the European Space Agency's Copernicus Contributing Missions (CCM) project.

%
%
\bibliographystyle{splncs04}
\bibliography{main}
\end{document}